\documentclass[letterpaper, 10 pt, conference]{ieeeconf}  

\IEEEoverridecommandlockouts                              

\usepackage{times}
\usepackage[pdftex]{graphicx}
\usepackage[ruled,linesnumbered]{algorithm2e}

\makeatletter
\newcommand{\removelatexerror}{\let\@latex@error\@gobble}
\newcommand{\titlefigurecaption}[2]{%
    \refstepcounter{figure}%
    \label{#1}%
    \@makecaption{\fnum@figure}{#2}%
}
\makeatother

\usepackage{subfigure}
\usepackage{amsmath,amssymb,amsopn,amstext,amsfonts}
\usepackage{cancel}
\usepackage[space]{cite}
\usepackage{balance}
\usepackage{color}
\usepackage{mathtools}
\usepackage{bm}

\usepackage{diagbox}
\usepackage{float}
\usepackage{placeins}
\usepackage{epstopdf}
\usepackage{pifont}
\usepackage{fixltx2e}
\usepackage{amsmath}
\usepackage{multirow}
\usepackage{booktabs}
\usepackage{url}
\usepackage{verbatim}
\usepackage[linkcolor=black,citecolor=black,urlcolor=black,colorlinks=true]{hyperref}
\usepackage{soul,xcolor}
\usepackage{overpic}
\usepackage{bbding}
\usepackage{amsmath,amssymb}
\usepackage{makecell}
\usepackage{CJKutf8}

\graphicspath{{figures/}}
\DeclareGraphicsExtensions{.png,.jpg,.eps,.pdf,.jpeg}

\newcommand{\methodname}{HERO: Hierarchical Region-Aware Reinforcement Learning for Autonomous Exploration}

\title{\LARGE \bf
\methodname
}

\IEEEoverridecommandlockouts
\author{%
    Xulin Xiao\textsuperscript{1}, Shengyi Liang\textsuperscript{1},
    Mingjie Zhang\textsuperscript{1}, Chengkai Wu\textsuperscript{1},
    Chen Chen\textsuperscript{4}, \\
    Zhigang Li\textsuperscript{4}, Jinni Zhou\textsuperscript{1},
    Jun Ma\textsuperscript{1}, Boyu Zhou\textsuperscript{2,3,\ensuremath{\dagger}}%
    \thanks{\textsuperscript{\ensuremath{\dagger}}\,\textbf{Corresponding Author}}%
    \thanks{\textsuperscript{1}\,The Hong Kong University of Science and Technology (Guangzhou), Guangzhou, China.}%
    \thanks{\textsuperscript{2}\,Southern University of Science and Technology, Shenzhen, China.}%
    \thanks{\textsuperscript{3}\,Differential Robotics, Hangzhou, China.}%
    \thanks{\textsuperscript{4}\,Huawei Technologies Co., Ltd., Shenzhen, China.}%
    \thanks{\texttt{zhouby@sustech.edu.cn}}%
}

\IEEEaftertitletext{%
    \vspace{0.5em}%
    \begin{minipage}{\textwidth}%
        \centering
        \includegraphics[width=\textwidth]{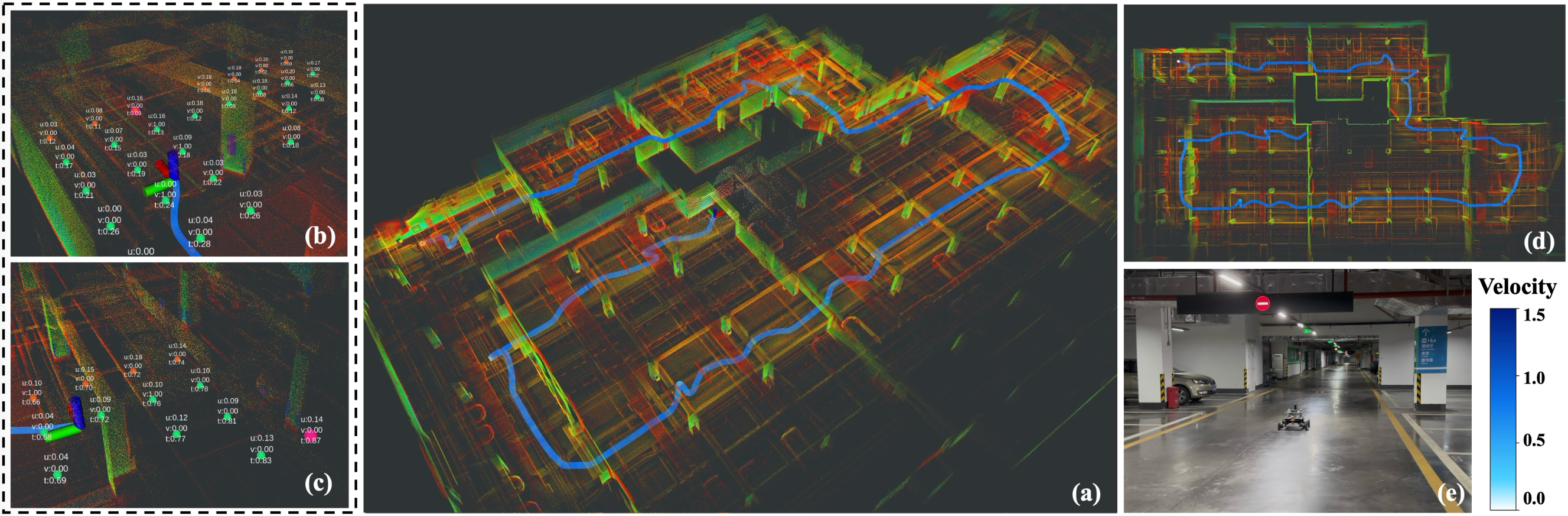}%
        \titlefigurecaption{fig:main}{Autonomous exploration of an underground parking garage (Site~I) using HERO. (a) Reconstructed point-cloud map and executed trajectory. (b) and (c) Local candidate-node evaluations during exploration. (d) Top view of the map and trajectory. (e) Robot in the test environment. Trajectory color indicates robot velocity in m/s.}%
    \end{minipage}%
    \vspace{0.5em}%
}

\begin{document}
\begin{CJK*}{UTF8}{gbsn}

\maketitle
\thispagestyle{empty}
\pagestyle{empty}

\begin{abstract}
Autonomous exploration is widely used to map unknown environments, yet maintaining both path efficiency and low online computation remains difficult as environment scale increases. Hierarchical planners reduce global search through coarse regional representations but often rely on handcrafted local viewpoint-selection rules. Graph-based reinforcement learning can improve local decisions, but its inference cost grows with the exploration graph and global direction is usually represented only implicitly. To address these limitations, we present HERO, a hierarchical framework that assigns distinct responsibilities to global planning and learned local decision-making, enabling each level to focus on the function it is best suited to perform. To reduce the growing inference cost of graph-based policies, HERO restricts learned target selection to reachable candidates within the current region while using region-level planning to provide long-horizon guidance. To avoid informative targets causing excessive deviation from this global direction, the proposed Dual-Branch Graph Policy balances Frontier Utility with guidance toward the next region, improving both local exploration and global progress. Experiments on unseen 2D maps show the lowest overall mean path length and target-decision cost among the evaluated methods. MARSIM simulations and deployments at three real-world sites further demonstrate the framework's applicability with LiDAR-based mapping and chassis dynamics.
\end{abstract}




\section{Introduction}
\label{sec:intro}

Autonomous exploration is essential for robots operating in unknown environments, underpinning applications like search-and-rescue and industrial inspection. The fundamental goal is to maximize map coverage while minimizing travel cost. However, achieving efficient exploration in large-scale environments remains a significant challenge. Extensive research therefore continues to investigate how to make effective exploration decisions with low computational overhead.

Despite extensive research, both classical and learning-based exploration methods face fundamental bottlenecks. Classical methods can provide deterministic, long-horizon global path that directs exploration toward complete coverage. However, their handcrafted rules struggle to adapt to diverse, complex environments. Conversely, reinforcement learning (RL) enables environment-adaptive target selection through data-driven training, which is effective within small, bounded regions. Yet as environment scale grows, exploration becomes a long-horizon task: rewards are sparse and delayed, each episode spans hundreds of sequential decisions before coverage completes, and the exploration graph grows continuously with the mapped area, making both training convergence and online inference increasingly costly. Without explicit global direction, RL decisions in unseen environments can drift from the overall coverage objective, accumulating unnecessary detours.

To resolve these complementary limitations, we present HERO, a hierarchical framework that establishes a novel division of labor between structured global planning and learned local decision-making. Our design assigns each layer the role it excels at. At the global level, an explicit region-level coverage plan provides long-horizon global guidance that end-to-end RL struggles to learn. At the local level, an RL policy performs environment-adaptive target selection within a bounded candidate graph, offering the adaptive flexibility that handcrafted rules lack. Restricting the graph locally keeps the policy inference cost bounded as the explored map grows. Meanwhile, the explicit global order relieves the policy from learning the entire exploration sequence, enabling it to focus strictly on balancing local frontier utility against global progress.

Realizing this hierarchy, however, is not straightforward. The first challenge is a gap between the two levels: the global planner specifies which region to visit next, but this region-level intent does not directly translate into actionable local decisions. The local policy must select among individual candidate nodes, yet it lacks explicit knowledge of how each candidate actually connects to the target region. We address this by introducing the Guidance-Distance Feature (GDF), which converts the region-level visitation order into a candidate-level path signal that the policy can utilize directly. The second challenge concerns how these two sources of information should be represented. We observe that local exploration value and global progress exhibit distinct, potentially conflicting objectives; a shared representation risks allowing one to dominate the other, degrading either coverage or path efficiency. This observation motivates our Dual-Branch Graph Policy, which encodes the two signals independently before fusing them, preserving their distinct semantics while enabling the policy to balance information gain against global progress at decision time.

We evaluate HERO in comparative experiments and ablation studies, assessing environment coverage, path efficiency, and online decision-making efficiency. The results show that our method maintains high coverage in largescale unknown environments while reducing unproductive detours and the computational costs. By deriving GDF, which represents global guidance, and combining it with local frontier utility through a Dual-Branch Graph Policy, the framework connects explicit global planning with learned local decision making.   It thus offers a new approach to autonomous exploration tasks that require both global task prioritization and flexible local decisions. Our main contributions are as follows:

\begin{enumerate}
    \item We propose a hierarchical framework that assigns structured global planning and RL-based exploration policy their complementary roles, providing long-horizon guidance while retaining environment-adaptive flexibility within a bounded decision scope.

    \item We introduce a Dual-Branch Graph Policy, equipped with GDF as explicit guidance, that learns to balance local frontier information against progress along the global path.

    \item We conduct comprehensive experiments in simulation and the real world to demonstrate the feasibility, stability, and real-time performance of the proposed method.
\end{enumerate}

\begin{figure*}[t]
    \centering
    \vspace*{5pt}
    \includegraphics[width=\textwidth]{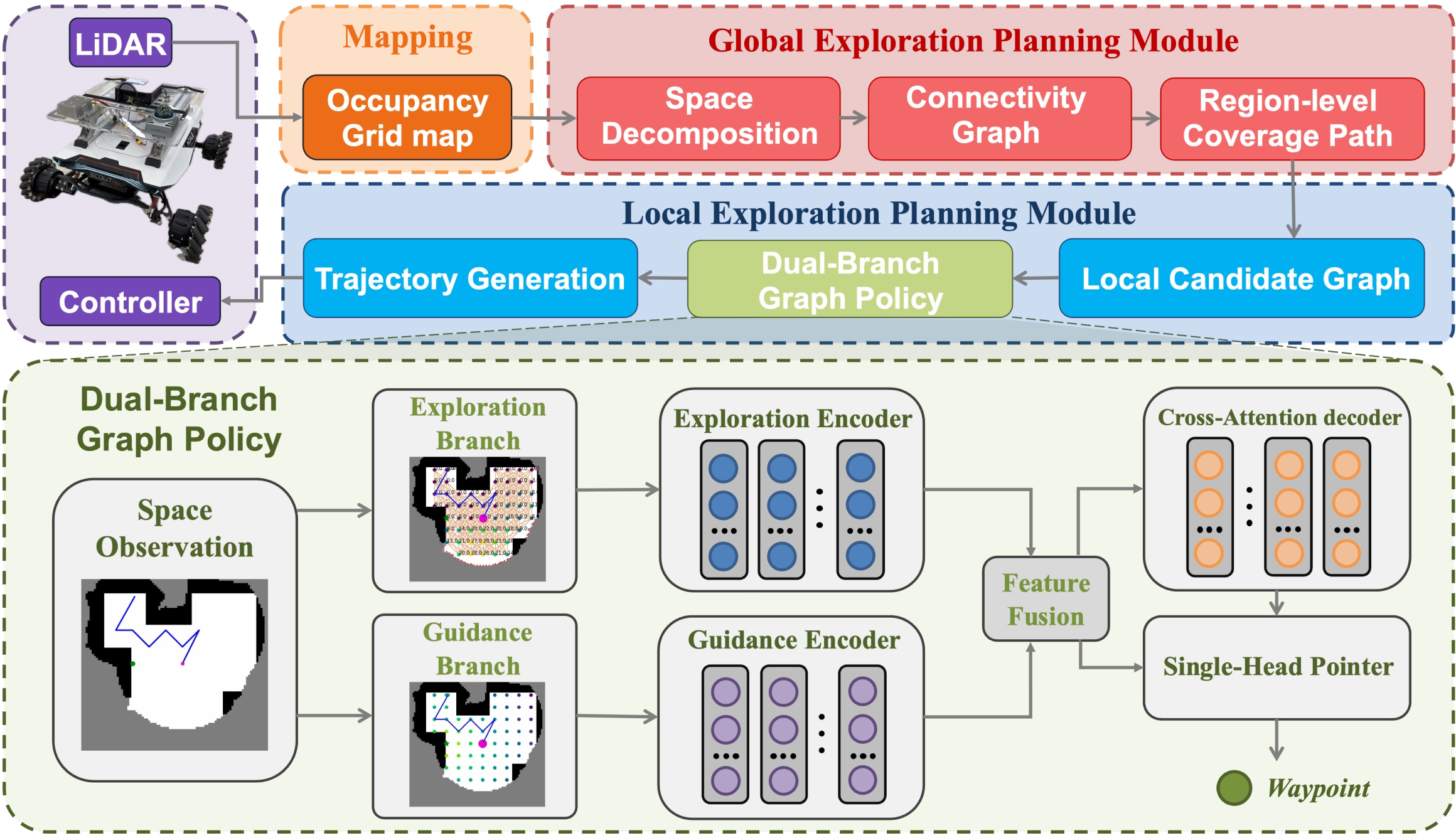}
    \caption{Overview of HERO. \emph{Top:} LiDAR observations update the occupancy grid map. The global module determines the order of region visits and sets a guide point in the next region, while the local module builds a graph of reachable candidate positions, selects a target, and executes a trajectory. \emph{Bottom:} The Dual-Branch Graph Policy separately encodes Frontier Utility and the Guidance-Distance Feature, fuses the resulting representations through cross-attention, and selects a target from the feasible neighbors.}
    \label{fig:system_overview}
    \vspace{-0.4cm}
\end{figure*}

\section{Related Work}
\label{sec:related}

\subsection{Local Exploration Planning}

Classical frontier-based exploration selects targets from frontiers detected in the current map~\cite{yamauchi1997frontier}. Later approaches score candidate viewpoints based on expected visible area and travel distance to balance information gain against path cost~\cite{gonzalez2002navigation}, while information-theoretic methods evaluate candidate actions using expected map information and localization uncertainty~\cite{bourgault2002information}. Receding-horizon methods evaluate sampled viewpoints or short path branches and repeat local planning as the map changes~\cite{bircher2016receding}. Their strength is to use observations for explicit, online selection among local exploration targets.

However, candidate generation and evaluation remain largely hand-designed, so the trade-off between information gain and travel cost may not readily adapt to different environment structures. Explicitly evaluating more candidates or sampled path branches can also increase per-cycle computation. Balancing candidate coverage and environment-adaptive local decisions under a limited computation budget therefore remains a challenge.

\subsection{Global Exploration Planning}

Hierarchical methods separate global planning from local execution, using sparse global representations for long-horizon paths and detailed maps locally~\cite{poncela2002efficient}. FUEL orders frontier regions globally and refines viewpoints and paths locally~\cite{zhou2021fuel}. TARE uses a locally dense but globally sparse representation for detailed local and coarse global planning~\cite{cao2021tare}. FALCON uses online regional connectivity to plan a coverage path and incorporates its visitation intent into local optimization~\cite{zhang2024falcon}. These methods give long-horizon exploration an explicit planning structure while reserving detailed computation for the local horizon.

However, local candidate selection in most hierarchical exploration methods still relies mainly on handcrafted objectives, sampling-based planning, or deterministic optimization. More recently, HEADER introduces learned local decision-making over a hierarchical graph and injects global guidance through binary guidepost features indicating whether a node lies on the global reference path~\cite{cao2025header}. While this design incorporates global path information into the local policy, binary path membership does not directly quantify how much each candidate viewpoint advances toward the current global target. The missing link is therefore not whether global information can reach the local policy, but how to translate it into fine-grained, learnable guidance over all candidate viewpoints. Such candidate-level guidance would let a policy balance information acquisition against global progress on a bounded local graph without learning the map-wide visitation order.

\subsection{Learning-Based Exploration}

Learning-based methods learn target-selection policies from data or robot--environment interactions, allowing local decisions to respond to the observed state without relying solely on handcrafted scores. Chen et al. use graph neural networks to select sensing actions from a graph of exploration information~\cite{chen2020autonomous}. ARiADNE uses attention and RL to learn the trade-off between information acquisition and travel cost~\cite{cao2023ariadne}, while graph sparsification and privileged learning improve large-scale inference efficiency~\cite{cao2024deep}. Occupancy anticipation predicts unobserved space from partial maps~\cite{ramakrishnan2020occupancy}, and uncertainty-aware planning uses uncertainty in predicted maps to evaluate candidate paths~\cite{georgakis2022uncertainty}. CogniPlan combines multiple global-layout predictions with graph attention~\cite{wang2025cogniplan}.

Although learning makes local decisions more flexible, graph policies such as ARiADNE learn long-horizon target selection from graph observations and training rewards rather than using an explicit online region order~\cite{cao2023ariadne,cao2024deep}. Map prediction can guide candidate paths using inferred global structure, but its added layout-inference stage has implementation-dependent online cost. Explicit global ordering and learned local selection thus offer complementary strengths, yet conveying the online region order to individual candidates without expanding the local decision graph remains an open design question.

\section{Problem Formulation and System Overview}
\label{sec:problem_formulation}
\label{sec:system_overview}

The problem considered in this paper is exploring an unknown and bounded 2D environment 
$V \subset \mathbb{R}^2$ using a mobile robot and building an occupancy grid map of the accessible space 
$V_{\mathrm{acc}} \subset V$ from LiDAR observations. 
The map $\mathcal{M}$ is partitioned into free, occupied, and unknown cell sets, denoted by 
$\mathcal{M}_{\mathrm{free}}$, $\mathcal{M}_{\mathrm{occupied}}$, and $\mathcal{M}_{\mathrm{unknown}}$, respectively, with frontiers defined at the free--unknown boundary. 
The exploration planner should provide collision-free paths through reachable targets, along which the robot covers 
$V_{\mathrm{acc}}$ with minimal travel distance and online target-decision cost.

An overview of the proposed exploration framework is shown in Fig.~\ref{fig:system_overview}. We propose our hierarchical exploration framework to achieve lightweight computation while inheriting the merits of a learning-based local planner. At the high level, the Topology-Guided Global Exploration Planning module generates the Region-level Coverage Path (RCP), an ordered sequence of regions that provides global exploration guidance (Sec.~\ref{subs:Topology-guided Global Coverage Planning}). To bridge this global guidance and local target selection, we construct a Guide-Point-Conditioned Local Candidate Graph, hereafter the Local Candidate Graph (LCG), encoding local exploration candidates (Sec.~\ref{subs:Guide-Point-Conditioned Local Candidate Graph}). Leveraging this graph representation, we further devise a novel Dual-Branch Graph Policy Learning module for next-target selection (Sec.~\ref{subs:Dual-Branch Graph Policy Learning}). The overall method jointly considers global exploration progress while preserving local exploration coverage, thereby maximizing overall exploration efficiency.

\section{Methodology}
\label{sec:methodology}


\subsection{Topology-Guided Global Exploration Planning}
\label{subs:Topology-guided Global Coverage Planning}
Without explicit guidance on the global exploration direction, a local policy makes decisions based only on a limited set of candidates. We therefore plan global coverage on a Region Connectivity Graph constructed from the occupancy grid, without requiring the local policy to encode the complete map. Specifically, we first divide the map into uniform grid blocks. Within each block, connected cells with the same occupancy state are merged into compact regions. The centroid of each region is extracted as a graph node to provide continuous global guidance. A topological edge is inserted between adjacent regions if a feasible path can be found to connect their centroids.

We evaluate connectivity with unknown-aware A*~\cite{hart1968formal}, which permits traversal through unknown cells because they may connect regions not yet observed, but assigns them a higher cost than known free cells. The one-step cost on \(\mathcal{M}_t\) is
\[
c_t(q,q')=
d(q,q')
\begin{cases}
1, & q'\in\mathcal{M}_{t,\mathrm{free}},\\
\lambda_u, & q'\in\mathcal{M}_{t,\mathrm{unknown}},\\
+\infty, & q'\in\mathcal{M}_{t,\mathrm{occupied}},
\end{cases}
\]
where \(d(q,q')\) is the distance between adjacent cells and \(\lambda_u>1\) is the cost multiplier for unknown cells.
Using graph-path costs between regions, the global planner solves an open asymmetric traveling-salesman problem with LKH to obtain the RCP \(\Pi_t=(z_{i_1},z_{i_2},\ldots,z_{i_K})\). Regional connectivity and the remaining visitation order are updated as the map changes.
The center of the next region \(z_{i_1}\) defines the current guide point \(g_t\), exposing the next global direction.

\subsection{Guide-Point-Conditioned Local Candidate Graph}
\label{subs:Guide-Point-Conditioned Local Candidate Graph}

\begin{figure}[!t]
    \centering
    \vspace*{5pt}
    \includegraphics[width=\columnwidth]{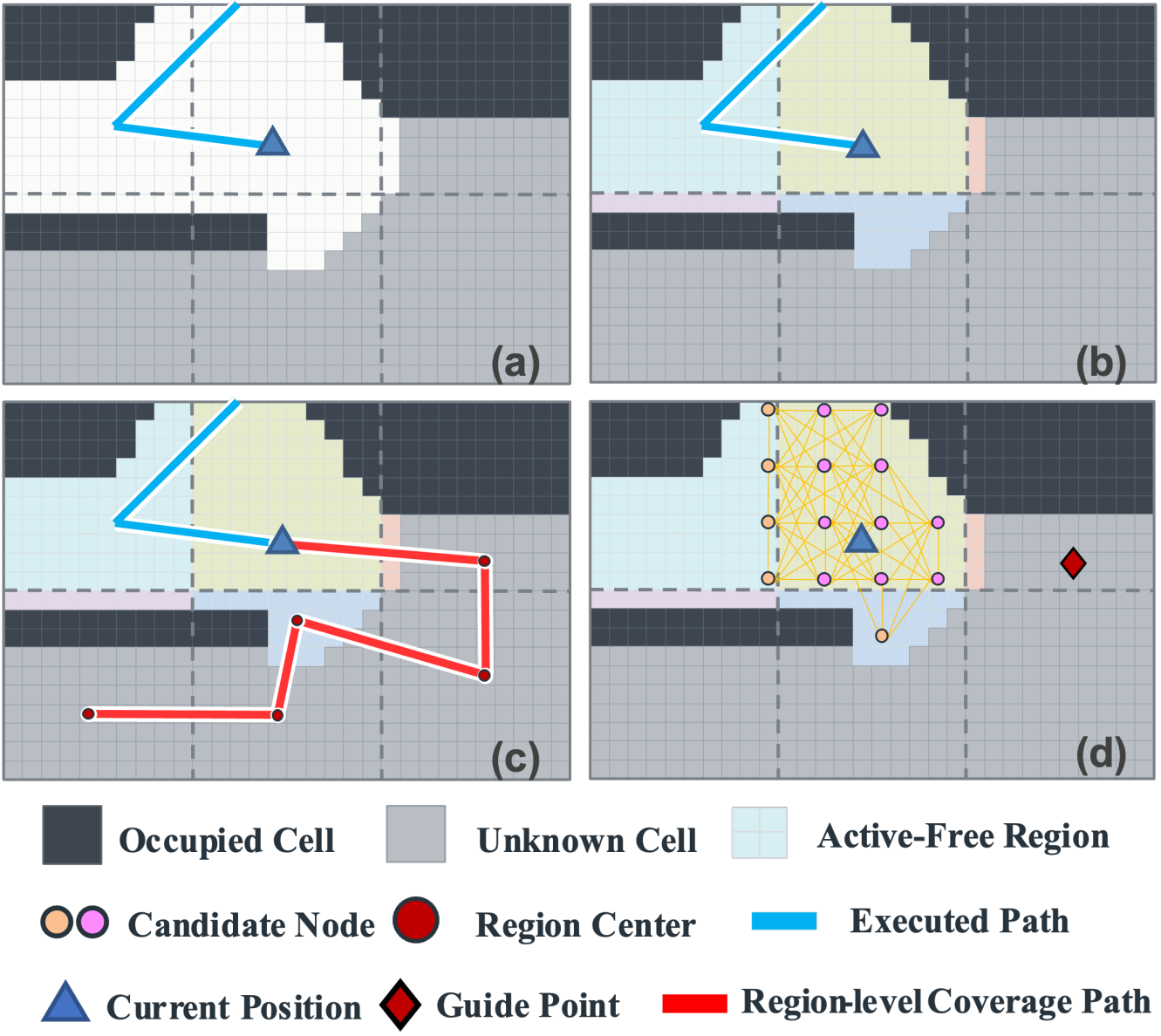}
    \caption{Construction of the Guide-Point-Conditioned Local Candidate Graph. (a) Fixed-size blocks partition the online occupancy map. (b) Connected regions are extracted within each block. (c) The Region-level Coverage Path orders region centers; the next region defines the guide point. (d) Collision-free connections among current-region candidates and one transition layer in adjacent regions.}
    \label{fig:local_candidate_graph_construction}
    \vspace{-0.4cm}
\end{figure}

A global exploration direction does not directly translate into candidate-level preferences, because the guide point does not specify a concrete local target. The local policy must therefore evaluate reachable candidates by balancing Frontier Utility against guidance cost. To deal with this problem, we construct the LCG, which additionally retains a layer of reachable candidate nodes in adjacent regions that are directly connected to the current region, thereby providing transitional connections for inter-region motion (Fig.~\ref{fig:local_candidate_graph_construction}).

We sample candidate nodes on a regular grid within the robot's reachable space, retaining nodes in the current region and a single transition layer of their directly connected exterior neighbors. Their union forms \(\mathcal{V}_t^L\), and node \(v_i\) has position \(\mathbf{p}_i=\mathbf{p}(v_i)\). Applying the same collision-free adjacency rule to current-region and transition nodes yields the edge set \(\mathcal{E}_t^L\) and the LCG \(\mathcal{G}_t^L=(\mathcal{V}_t^L,\mathcal{E}_t^L)\).
Let \(v_t\) denote the robot's current node, such that \(\mathbf{p}(v_t)=\mathbf{p}_t\). The policy selects one of its neighbors as the next local target; this neighbor may be a transition node, and its position is denoted by \(\mathbf{p}_{t+1}\). After execution, the LCG is updated for the robot's current region. Fixed block size and grid spacing keep the policy input bounded.

Each node \(v_i\) is described by its normalized coordinates \((\bar{p}_i^x,\bar{p}_i^y)\) relative to the current position, normalized Frontier Utility \(\tilde{u}_i\), visitation flag \(b_i\), and Guidance-Distance Feature \(h_i\). The relative coordinates are defined as
\[
[\bar{p}_i^x,\bar{p}_i^y]
=
\frac{\mathbf{p}_i-\mathbf{p}_t}{S_L},
\]
where \(S_L\) is the local-update scale. Relative coordinates represent the same structure consistently across map locations.
The raw Frontier Utility \(u_i\), defined as the number of frontiers observable from \(v_i\), is normalized across maps using the reference utility \(u_{\mathrm{ref}}\) and scale parameter \(\beta\):
\[
\tilde{u}_i=
\tanh\left(
\beta\frac{u_i}{u_{\mathrm{ref}}}
\right).
\]
The visitation flag \(b_i\in\{0,1\}\) distinguishes new candidates from previously visited nodes.
The guide-point position alone does not describe how obstacles affect each candidate's route to it. We therefore represent candidate-level guidance using path distance on the current occupancy map. Let \(D_t(\mathbf{p},g_t)\) denote the unknown-aware A* path distance from position \(\mathbf{p}\) to \(g_t\). The GDF \(h_i\) normalizes this distance as
\[
h_i=
1-
\frac{D_t(\mathbf{p}_i,g_t)}
{\displaystyle\max_{j:D_t(\mathbf{p}_j,g_t)<\infty}D_t(\mathbf{p}_j,g_t)}.
\]
GDF may use unknown cells to express guidance before a fully observed route exists. A larger \(h_i\) indicates a smaller remaining guide-point cost relative to the candidate set. Thus, \(\tilde{u}_i\) and \(h_i\) characterize local exploration value and progress toward the guide point, respectively, and may provide different assessments of the same candidate.

\subsection{Dual-Branch Graph Policy Learning}
\label{subs:Dual-Branch Graph Policy Learning}

\subsubsection{Dual-Branch Graph Policy}

The LCG provides both exploration and guidance information for candidate nodes, but these signals serve distinct and potentially conflicting objectives. Encoding them in a shared representation may obscure their individual roles in local coverage and global progress. We therefore adopt a Dual-Branch Graph Policy with two independent graph-attention encoders~\cite{velivckovic2017graph}, allowing the two signals to be learned separately before being balanced for target selection.

The exploration and guidance branches receive \(\mathbf{x}^{\mathrm{exp}}_i=[\bar{p}_i^x,\bar{p}_i^y,\tilde{u}_i,b_i]\) and \(\mathbf{x}^{\mathrm{gui}}_i=[\bar{p}_i^x,\bar{p}_i^y,b_i,h_i]\), respectively. The fused node features are decoded with respect to the current node through cross-attention~\cite{vaswani2017attention}. A pointer mechanism~\cite{vinyals2015pointer} then scores the feasible neighboring nodes and selects the highest-scoring one as the next local target. The detailed network architecture is illustrated in Fig.~\ref{fig:system_overview}.

\subsubsection{Reward and Policy Optimization}

A reward based solely on newly explored areas cannot account for travel cost, inter-region guidance, or repeated visits. We therefore train the local policy with an information-gain reward combined with travel cost, revisit penalty, and Guide-Point Regret, which penalizes travel that does not contribute to progress toward the guide point.

Let \(\Delta f_t\) be the number of newly observed frontiers, \(l_t\) the action length, and \(n_t\) the selected candidate's visit count; we use the path distance \(D_t(\mathbf{p},g_t)\) defined in Sec.~\ref{subs:Guide-Point-Conditioned Local Candidate Graph}. Guide-Point Regret \(\Delta_t^g\) and the dense reward are defined as
\[
\begin{aligned}
\Delta_t^{g}={}&l_t+D_t(\mathbf{p}_{t+1},g_t)-D_t(\mathbf{p}_t,g_t),\\
r_t^{\mathrm{dense}}={}&
\alpha_f\widetilde{\Delta f}_t
-\alpha_d\frac{l_t}{S_L} \\
&-\alpha_g\Delta_t^{g}
-\alpha_v\min\{\max(n_t-1,0),n_{\max}\},
\end{aligned}
\]
GDF represents the candidate-to-guide path cost in the policy input, while Guide-Point Regret evaluates the executed action in the reward. The frontier increment \(\widetilde{\Delta f}_t\) uses the same hyperbolic-tangent normalization as node Frontier Utility. The weights \(\alpha_f\), \(\alpha_d\), \(\alpha_g\), and \(\alpha_v\) control the corresponding reward terms, and \(n_{\max}\) caps the revisit count.

The fixed exit \(e\) defines the endpoint. Successful termination requires reaching \(e\) after either the coverage ratio \(\rho_t\) reaches \(\rho_{\star}\) or every local candidate has zero Frontier Utility. The evacuation-permission and exit-reaching indicators are
\[
\begin{aligned}
\chi_t&=\mathbb{I}\!\left[
\rho_t\ge\rho_{\star}
\;\lor\;
\max_{v_i\in\mathcal{V}_t^L}u_i=0
\right],\\
\zeta_t&=\mathbb{I}\!\left[
d_{\mathrm{gt}}(\mathbf{p}_{t+1},e)\le\epsilon_e
\right].
\end{aligned}
\]
The corresponding exit reward is
\[
r_t^{\mathrm{exit}}=
\begin{cases}
R_{+}, & \zeta_t=1\ \text{and}\ \chi_t=1,\\
-R_{-}, & \zeta_t=1\ \text{and}\ \chi_t=0,\\
0, & \zeta_t=0.
\end{cases}
\]
Here, \(d_{\mathrm{gt}}\) is the ground-truth shortest-path distance, \(\epsilon_e\) the exit tolerance, and \(R_{+}\) and \(R_{-}\) the successful-evacuation reward and early-arrival penalty. Early arrival does not terminate the episode, and unsuccessful trajectories are truncated at the maximum episode length. The total reward is \(r_t=r_t^{\mathrm{dense}}+r_t^{\mathrm{exit}}\).

We train the policy with proximal policy optimization (PPO)~\cite{schulman2017proximal} and generalized advantage estimation (GAE)~\cite{schulman2015high}. An independently parameterized critic follows the dual-branch encoding pattern, estimates \(Q_\phi(s_t,j)\) for each valid candidate, and uses their policy-weighted mean as the state value.
GAE uses discount factor \(\gamma\) and parameter \(\lambda\), followed by advantage normalization. The PPO clipping range and entropy coefficient are \(\epsilon\) and \(\eta\); the critic minimizes the squared error between the selected action value and its return target.

\subsubsection{Training-Scenario Curriculum}

Partial observations make the regional order and guide point uncertain early in exploration; guidance stabilizes as coverage increases. We model this transition during training by progressively concentrating boundary guide-point samples near a fixed exit. Each online-generated scenario contains a polygonal map \(\mathcal{M}\), an initial position \(\mathbf{p}_0\) sampled uniformly from free space, and an exit \(e\). We sample \(e\) uniformly from robot-reachable exterior-contour nodes on the sampling lattice, subject to
\[
d_{\min}\le d_{\mathrm{gt}}(\mathbf{p}_0,e)<\infty,
\]
where \(d_{\min}\) keeps the initial position away from the exit.
Let \(\mathcal{B}=\{\mathbf{b}_i\}\) be the set of exterior-contour free cells reachable from the initial position, \(\delta_i=\|\mathbf{b}_i-e\|_2\), and \(\bar{\delta}_i=\delta_i/\max\{\max_j\delta_j,s_c\}\), where \(s_c\) is the grid resolution. The coverage ratio \(\rho\) is mapped to a contraction coefficient by a normalized hyperbolic tangent:
\[
\kappa(\rho)=
\frac{
\tanh\!\left((\rho-\rho_c)/w_c\right)-\tanh\!\left(-\rho_c/w_c\right)
}{
\tanh\!\left((1-\rho_c)/w_c\right)-\tanh\!\left(-\rho_c/w_c\right)
},
\]
where \(\rho_c\) and \(w_c\) control the center and width of the contraction schedule, respectively. While \(\rho_t<\rho_{\star}\), the current guide point is resampled after each waypoint according to
\[
\Pr(g_t=\mathbf{b}_i\mid\rho_t)=
\frac{
\exp\!\left[-\tau\kappa(\rho_t)\bar{\delta}_i^2\right]
}{
\sum_{\mathbf{b}_j\in\mathcal{B}}
\exp\!\left[-\tau\kappa(\rho_t)\bar{\delta}_j^2\right]
},
\]
where \(\tau\) controls concentration toward the exit and \(\kappa(0)=0\) yields initially uniform contour sampling. Once \(\rho_t\ge\rho_{\star}\), we set \(g_t=e\) for final evacuation. Using the same \(g_t\) for GDF and Guide-Point Regret keeps guidance consistent between observation and reward.

\subsubsection{Implementation Details}

We use a \(0.5\,\mathrm{m}\)-resolution occupancy grid partitioned into \(85\times85\)-cell blocks. For LCG construction, \(S_L=80\,\mathrm{m}\), candidate nodes are sampled every \(4\,\mathrm{m}\), and collision-free edges use a \(5\times5\) lattice stencil. We set \(\lambda_u=2.5\); during training, \(\beta=\ln(3)\) and \(u_{\mathrm{ref}}=80\).

The two branches use independent 6-layer, 8-head graph-attention encoders with 128-dimensional embeddings. Their fused outputs are passed to a single-layer decoder for action selection.

Training uses randomly generated maps smaller than \(50\,\mathrm{m}\times50\,\mathrm{m}\), with at most 128 actions per episode. We set \(\rho_\star=0.99\), \(d_{\min}=10\,\mathrm{m}\), and \(\epsilon_e=0.5\,\mathrm{m}\). The reward parameters are \((\alpha_f,\alpha_d,\alpha_g,\alpha_v)=(1,5,0.05,0.10)\), \(n_{\max}=8\), and \((R_+,R_-)=(10,2)\); the curriculum parameters are \((\rho_c,w_c,\tau)=(0.5,0.2,12)\).

We train the policy with PPO and Adam on a single NVIDIA GeForce RTX 4080 SUPER GPU for approximately 2 hours, using \((\gamma,\lambda,\epsilon,\eta)=(0.99,0.95,0.12,0.004)\) and a learning rate of \(6\times10^{-6}\). Each update uses 3000 rollout samples, 6 optimization epochs, and a minibatch size of 32.

\section{Experiments}
\label{sec:experiments}

We evaluate HERO through 2D comparisons and ablations, MARSIM simulations~\cite{kong2023marsim}, and real-world deployments using the mobile-base planner adopted by Wu et al.~\cite{wu2024real,wu2026synergizing}.

\subsection{Comparison in 2D Simulated Environments}
\label{subs:2d_benchmark}

The 2D benchmark comprises 50 structured Room maps and 50 random Forest maps, each \(125\,\mathrm{m}\times125\,\mathrm{m}\) and unseen during training. Reaching 99.5\% coverage marks task completion; otherwise, a run ends after 3000 environment motion steps. Coverage and traveled distance at termination are averaged over all runs. Baselines include NBV~\cite{bircher2016receding}, ARiADNE~\cite{cao2023ariadne}, CogniPlan~\cite{wang2025cogniplan}, and FALCON~\cite{zhang2024falcon}. ARiADNE and CogniPlan use their released weights and configurations unchanged, with no Forest-specific retraining; all methods are timed on the same Intel Core i7-14700KF CPU.

Table~\ref{tab:2d_benchmark_results} shows that HERO completed the coverage task in both scene types and achieved the lowest overall mean traveled distance. Some baseline runs were truncated at the step limit because exploration stagnated or progressed too slowly; their recorded travel includes movement during stagnation. The clearest comparison with the closest hierarchical baseline is in Forest: HERO and FALCON both reached the target coverage on average, while HERO's mean traveled distance was 9.9\% lower. Since the two methods share the Topology-Guided Global Exploration Planning module, the shorter travel supports learned local target selection under regional guidance. CogniPlan traveled slightly less in Room. In Forest, HERO's mean traveled distance was 15.7\% lower than CogniPlan's, with higher mean final coverage.

\begin{figure}[!t]
    \centering
    \vspace*{5pt}
    \makebox[\columnwidth][l]{\includegraphics[width=1.0\columnwidth]{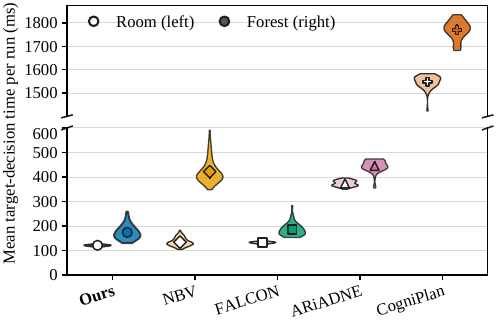}}
    \caption{Decision-module timing scores for 50 runs per scene: Room (left) and Forest (right). Hollow and filled markers show scene means; black segments indicate interquartile ranges. The axis omits the sample-free 600--1400~ms interval.}
    \label{fig:decision_time}
    \vspace{-0.6cm}
\end{figure}

The decision-time metric in Table~\ref{tab:2d_benchmark_results} and Fig.~\ref{fig:decision_time} sums per-run mean call times for the measured modules: global route optimization, local feature preparation, and inference for HERO; global and local route optimization for FALCON; graph/observation updates and inference for ARiADNE and CogniPlan; and planning for NBV.

HERO records the lowest scores in both scenes, 8.9\% and 6.2\% below FALCON in Room and Forest, respectively (Fig.~\ref{fig:decision_time}). Their global TSP times are similar, locating the advantage in the local stage: HERO's feature preparation and inference on a bounded graph are faster than FALCON's local sequential ordering. This advantage persists despite costlier observation construction in Forest. More numerous and dispersed Forest frontiers increase NBV's viewpoint-evaluation and path-search costs. ARiADNE's growing global graph increases the multi-layer attention workload, with inference dominating its measured cost. CogniPlan adds layout prediction and associated state updates, increasing its workload beyond global graph inference.

\begin{figure}[!b]
    \centering
    \includegraphics[width=0.95\columnwidth]{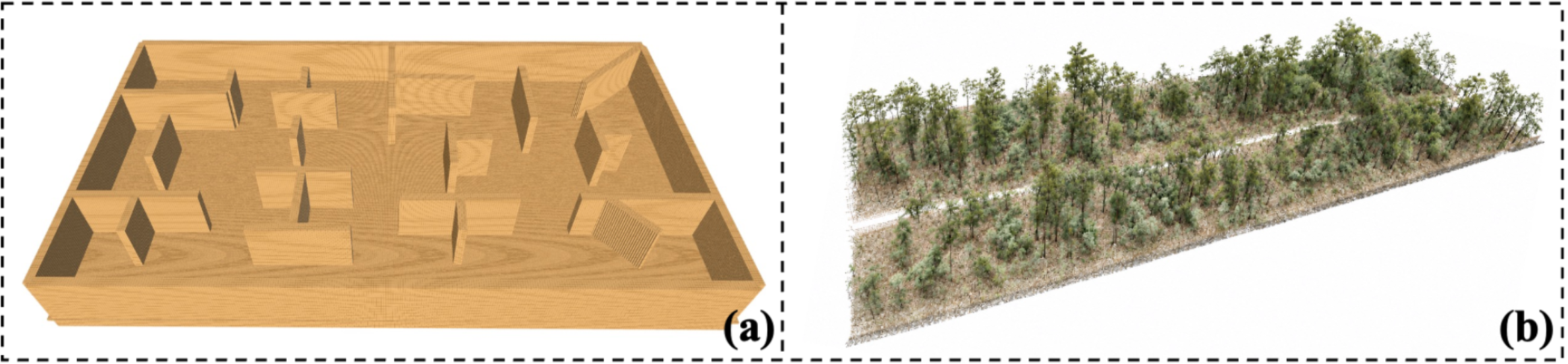}
    \caption{MARSIM scenes: (a) Office, $46.9~\mathrm{m} \times 90.9~\mathrm{m} \times 8.7~\mathrm{m}$; (b) Forest, $156.1~\mathrm{m} \times 47.9~\mathrm{m} \times 15.5~\mathrm{m}$.}
    \label{fig:marsim_scenes}
    \vspace{-0.2cm}
\end{figure}

\begin{figure*}[!t]
    \centering
    \vspace*{5pt}
    \includegraphics[width=\textwidth]{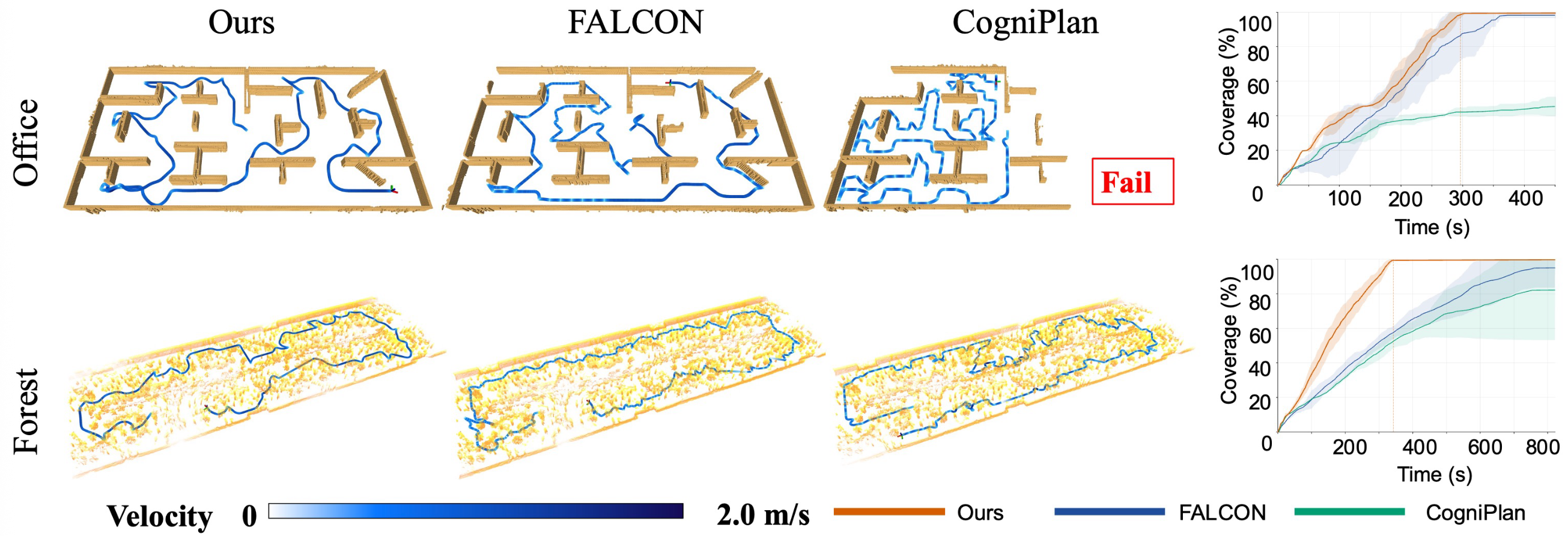}
    \caption{MARSIM results in Office (top) and Forest (bottom). The first three panels in each row show representative trajectories for HERO, FALCON, and CogniPlan, with color denoting velocity; the rightmost panel shows the mean normalized exploration progress and two-sided 95\% Student-\(t\) confidence interval over five runs.}
    \label{fig:marsim_results}
    \vspace{-8pt}
\end{figure*}

\begin{table}[!b]
    \caption{Mean results on the 2D benchmark.}
    \label{tab:2d_benchmark_results}
    \centering
    \footnotesize
    \setlength{\tabcolsep}{1.55pt}
    \renewcommand{\arraystretch}{1.1}
    \begin{tabular}{@{}llrrrr@{}}
        \toprule
        Method & Scene &
        \makecell{Coverage\\(\%)} &
        \makecell{Traveled\\distance (m)} &
        \makecell{Decision modules\\(ms)} &
        \makecell{Overall mean\\distance (m)} \\
        \midrule
        \multirow{2}{*}{NBV}
            & Room   & 99.531 & 1207.453 & 134.7   & \multirow{2}{*}{1433.527} \\
            & Forest & 99.251 & 1659.602 & 421.835 & \\
        \midrule
        \multirow{2}{*}{FALCON}
            & Room   & 97.212 & 663.651  & 133.324 & \multirow{2}{*}{898.261} \\
            & Forest & 99.521 & 1132.872 & 185.510 & \\
        \midrule
        \multirow{2}{*}{ARiADNE}
            & Room   & 96.482 & 590.766  & 373.743 & \multirow{2}{*}{1054.780} \\
            & Forest & 98.894 & 1518.794 & 445.064 & \\
        \midrule
        \multirow{2}{*}{CogniPlan}
            & Room   & 99.797 & \textbf{503.617} & 1548.174 & \multirow{2}{*}{857.137} \\
            & Forest & 98.191 & 1210.657 & 1769.265 & \\
        \midrule
        \multirow{2}{*}{\textbf{HERO}}
            & Room   & 99.558 & 509.435 & \textbf{121.424} & \multirow{2}{*}{\textbf{764.924}} \\
            & Forest & 99.518 & \textbf{1020.413} & \textbf{174.073} & \\
        \bottomrule
    \end{tabular}
    \vspace{3pt}

    \parbox{\columnwidth}{\scriptsize Coverage and distance means include all 50 runs per scene, including truncated runs; overall distance uses all 100 runs. Bold marks the best value per scene or overall.}
\end{table}

\subsection{Ablation Experiment}
\label{subs:ablation}

The ablation compares GDF input and encoding choices while retaining Topology-Guided Global Exploration Planning, the guide point, the LCG, and the guidance reward. \emph{Single w/o GDF} jointly encodes position, Frontier Utility, and visitation state; \emph{Single + GDF} adds GDF to the same encoder architecture. This pair tests explicit GDF input, while comparison with HERO assesses separate encoding of Frontier Utility and GDF before fusion. Without GDF, HERO's guidance branch would retain only position and visitation features shared with the exploration branch, losing its dedicated guidance input. The ablation therefore concerns input and encoding within the guidance framework, rather than removing global guidance. Every run contributes its attained coverage and traveled distance, including runs truncated at the step limit. All configurations use 50 runs per scene type.

\begin{table}[t]
    \caption{Mean ablation results in Room and Forest.}
    \label{tab:ablation_results}
    \centering
    \footnotesize
    \setlength{\tabcolsep}{3.0pt}
    \renewcommand{\arraystretch}{1.10}
    \begin{tabular*}{\columnwidth}{@{\extracolsep{\fill}}llrr@{}}
        \toprule
        Scene & Variant & Coverage (\%) & Distance (m) \\
        \midrule
        \multirow{3}{*}{Room}
        & \textbf{HERO}
        & \textbf{99.558}
        & \textbf{509.435} \\
        & Single + GDF
        & 76.076
        & 1479.627 \\
        & Single w/o GDF
        & 96.237
        & 912.119 \\
        \midrule
        \multirow{3}{*}{Forest}
        & \textbf{HERO}
        & \textbf{99.518}
        & \textbf{1020.413} \\
        & Single + GDF
        & 96.813
        & 1449.368 \\
        & Single w/o GDF
        & 95.622
        & 1534.058 \\
    \bottomrule
    \end{tabular*}
    \vspace{-9pt}
\end{table}

Table~\ref{tab:ablation_results} shows that \emph{Single + GDF} achieved lower mean coverage and substantially longer mean traveled distance than HERO in Forest, with some runs truncated before completion. The coverage loss was more pronounced in Room, accompanied by frequent cyclic target selections and exploration stagnation. The observed instability is consistent with interference between GDF and Frontier Utility in the shared encoder, as the two features can evaluate the same candidate differently.

In Room, \emph{Single w/o GDF} exhibited fewer cyclic target selections and step-limit terminations, with higher coverage and shorter travel than \emph{Single + GDF}. It nevertheless underperformed HERO in both scenes and \emph{Single + GDF} in Forest. Removing the explicit GDF input therefore alleviated unstable behavior in Room but also reduced the policy's access to candidate-level guidance.

These results suggest that GDF provides useful candidate-level guidance, but its benefit depends on the encoding design. HERO's stronger results in both scenes support separate encoding followed by fusion to combine exploration and guidance information. 

\begin{figure}[!t]
    \centering
    \includegraphics[width=\columnwidth]{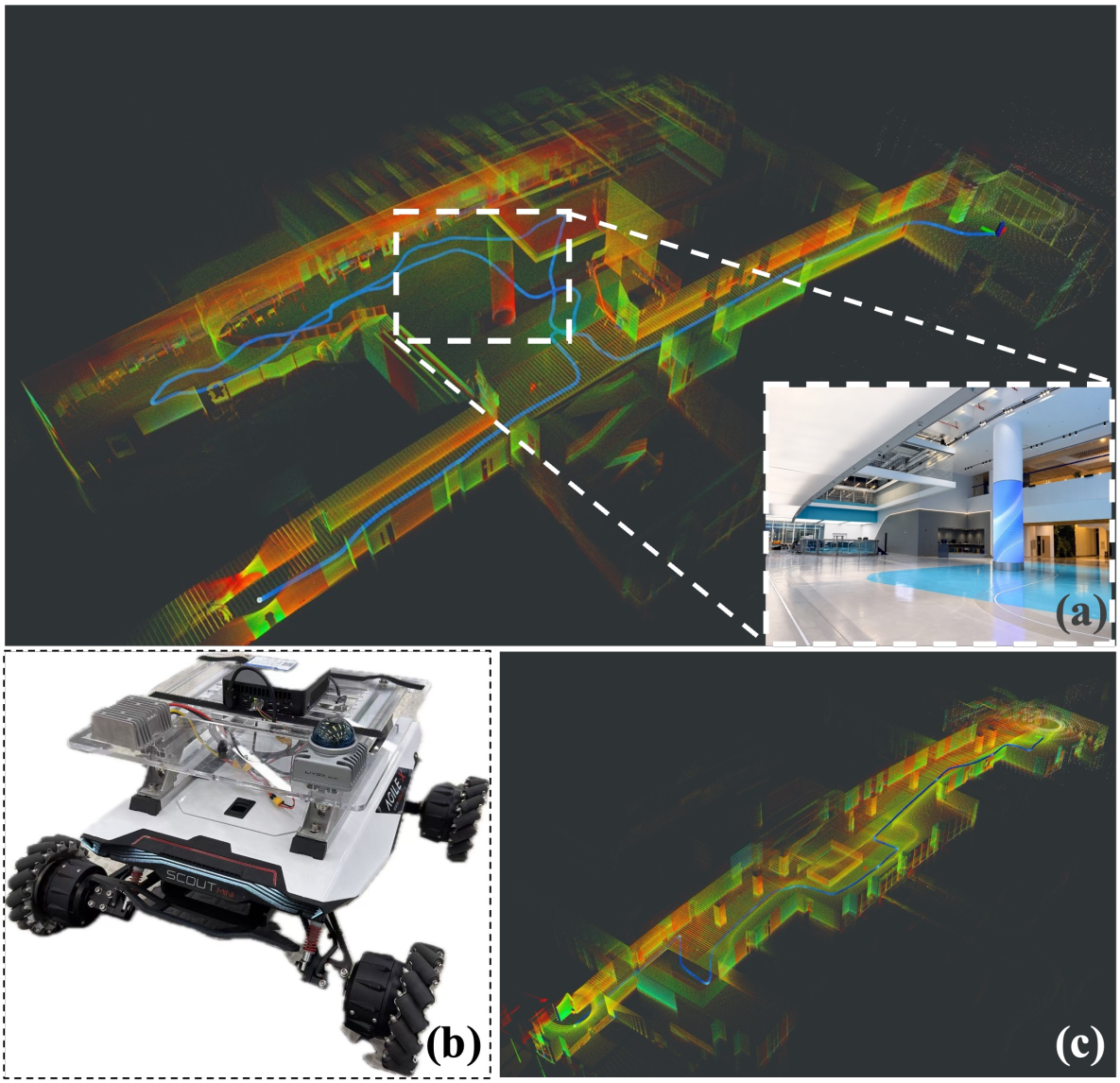}
    \vspace{-5pt}
    \caption{Real-world platform and exploration results. (a) Site~II hall; (b) SCOUT MINI platform; (c) Site~III corridor. Blue curves show the executed trajectories over the reconstructed point-cloud maps.}
    \label{fig:real_world_results}
    \vspace{-0.2cm}
    \vspace{-16pt}
\end{figure}

\subsection{Comparison in 3D Simulated Environments}
\label{subs:marsim}

\begin{table}[t]
    \caption{MARSIM results reported as mean \(\pm\) standard deviation over five runs.}
    \label{tab:marsim_single_run}
    \vspace{-4pt}
    \centering
    \footnotesize
    \renewcommand{\arraystretch}{1.08}
    \begin{tabular*}{\columnwidth}{@{\extracolsep{\fill}}llrr@{}}
        \toprule
        Scene & Method & Distance (m) & Time (s) \\
        \midrule
        \multirow{3}{*}{Office}
            & \textbf{HERO} & \(\mathbf{333.834\pm6.868}\) & \(\mathbf{280.397\pm12.202}\) \\
            & FALCON     & \(357.281\pm27.537\) & \(363.080\pm30.441\) \\
            & CogniPlan\(^{\dagger}\) & \(505.751\pm56.331\) & \(757.651\pm79.185\) \\
        \midrule
        \multirow{3}{*}{Forest}
            & \textbf{HERO} & \(\mathbf{403.863\pm6.595}\) & \(\mathbf{352.069\pm7.057}\) \\
            & FALCON     & \(466.495\pm46.520\) & \(622.972\pm71.500\) \\
            & CogniPlan  & \(580.451\pm140.855\) & \(756.706\pm189.365\) \\
        \bottomrule
    \end{tabular*}
    \vspace{3pt}

    \parbox{\columnwidth}{\scriptsize \(^{\dagger}\)Includes an incomplete run. Distance and time are measured until termination.}

    \vspace{-0.4cm}
\end{table}

The 3D simulations examine whether target-selection efficiency translates into faster exploration with online LiDAR mapping and chassis dynamics. HERO, FALCON, and CogniPlan each run five trials in Office and Forest (Fig.~\ref{fig:marsim_scenes}). Office contains walls and internal partitions, while Forest is an elongated outdoor environment with dense vegetation, testing route execution across different spatial structures.

At HERO's scene-specific mean completion time, its progress lead over the strongest competing curve was 12.172 percentage points in Office and 41.956 points in Forest. This matched-time comparison uses the five-run mean progress curves in Fig.~\ref{fig:marsim_results}. Table~\ref{tab:marsim_single_run} reports the mean distance and elapsed time with standard deviations over five runs.

The trajectory visualizations complement the mean progress curves by showing the spatial paths and speed profiles of the representative runs.

\subsection{Validation in the Real World}
\label{subs:real_world}

Real-world deployments evaluate operation with onboard sensing, computation, and motion execution in a garage (Site~I), a hall (Site~II), and a corridor (Site~III) (Figs.~\ref{fig:main} and~\ref{fig:real_world_results}). The SCOUT MINI mecanum-wheel platform carries a Livox Mid-360 LiDAR and a NUC13ANK computer.

At Site~I (\(107~\mathrm{m}\times53~\mathrm{m}\times4~\mathrm{m}\)), the robot traveled \(265.8~\mathrm{m}\) in \(301.71~\mathrm{s}\); at Site~II (\(83~\mathrm{m}\times36~\mathrm{m}\times10~\mathrm{m}\)), it traveled \(178.1~\mathrm{m}\) in \(204.61~\mathrm{s}\); and at Site~III (\(130~\mathrm{m}\times50~\mathrm{m}\times5~\mathrm{m}\)), it traveled \(116.06~\mathrm{m}\) in \(103.5~\mathrm{s}\).

The maps and trajectories show operation across distinct layouts: connected aisles in the garage, an open hall linked to surrounding passages, and a corridor with repeated turns. These examples demonstrate how movement between regions and local target execution are integrated on the physical platform.

Across all three sites, the same simulation-trained policy operated onboard without real-world retraining, integrating Topology-Guided Global Exploration Planning, online mapping, and mobile-base execution. The reconstructed maps and executed trajectories support onboard deployment of the complete framework and transfer of the learned local policy to real environments.

\section{Conclusions}
\label{sec:conclusions}

This paper presented HERO, a hierarchical framework that assigns distinct roles to Topology-Guided Global Exploration Planning and learned local target-selection. Its bounded LCG and Dual-Branch Graph Policy separately encode GDF and Frontier Utility to balance regional progress and local coverage. On unseen 2D maps, HERO achieved the lowest overall mean traveled distance and decision-module time among the evaluated methods; ablations supported explicit guidance and branch-wise feature separation. MARSIM results showed faster exploration than competing methods, and deployments at three real-world sites demonstrated onboard operation without real-world retraining. Current validation is limited to a ground robot; future work will extend HERO to aerial robots and other resource-constrained edge platforms.


\balance
\bibliography{refs}

\end{CJK*}

\end{document}